%% file: GLoCIM.tex
\documentclass[11pt]{article}
\usepackage{GLoCIM}
\usepackage{times}
\usepackage{latexsym}
\usepackage[T1]{fontenc}
\usepackage[utf8]{inputenc}
\usepackage{microtype}

\usepackage{braket}
\usepackage{subfigure}
\usepackage[export]{adjustbox}
\usepackage{microtype}
\usepackage{graphicx}
\usepackage{amsmath}

\usepackage{booktabs}
\usepackage{multirow}
\usepackage{balance}
\usepackage{amssymb} 
\usepackage{tabularx} 
\usepackage{array}  
\usepackage{tikz}
\usepackage{pgfplots}
\pgfplotsset{compat=1.18}
\usepackage{subcaption}
\usepackage{algorithm} 

\usepackage{soul, color, xcolor}
\usepackage{coling}

\usepackage{makecell}
\usepackage{algcompatible}

\usepackage{bm}       

\usepackage{fancyhdr}
\definecolor{mycolor5}{RGB}{216, 191, 216}  
\definecolor{mycolor6}{RGB}{255, 182, 193}  
\definecolor{mycolor7}{RGB}{173, 216, 230}  

\AtBeginDocument{
  \let\mathbb\relax
  \DeclareMathAlphabet{\mathbb}{U}{msb}{m}{n}
}
\title{GLoCIM: Global-view Long Chain Interest Modeling for news recommendation}
\author{\outauthor} 
\author{
  Zhen Yang$^{1}$, Wenhui Wang$^{1}$, Tao Qi$^{2}$, Peng Zhang$^{1}$,\\
  \textbf{Tianyun Zhang$^{1}$, Ru Zhang$^{1}$, Jianyi Liu$^{1}$, Yongfeng Huang$^{2,3}$}  \\
  $^1$School of Cyberspace Security, Beijing University of Posts and Telecommunications, 100876 Beijing, China \\
  $^2$Department of Electronic Engineering, Tsinghua University, 100084 Beijing, China\\
  $^3$Zhongguancun Laboratory, 100094 Beijing, China\\ 
  \texttt{\{yangzhenyz,edwin\_wang,zhangpengzp,zhangtianyun,zhangru,liujy\}@bupt.edu.cn} \\
  \texttt{taoqi.qt@gmail.com} \quad \texttt{yfhuang@tsinghua.edu.cn} \\
}

\date{}
\vspace{5cm} 
\begin{document}
\maketitle
\thispagestyle{fancy} 
\begin{abstract}
Accurately recommending candidate news articles to users has always been the core challenge of news recommendation system. News recommendations often require modeling of user interest to match candidate news. Recent efforts have primarily focused on extracting local subgraph information in a global click graph constructed by the clicked news sequence of all users. Howerer, the computational complexity of extracting global click graph information has hindered the ability to utilize far-reaching linkage which is hidden between two distant nodes in global click graph collaboratively among similar users. To overcome the problem above, we propose a \textbf{G}lobal-view \textbf{L}\textbf{o}ng \textbf{C}hain \textbf{I}nterests \textbf{M}odeling for news recommendation (\textbf{GLoCIM}), which combines neighbor interest with long chain  interest distilled from a global click graph, leveraging the collaboration among similar users to enhance news recommendation. We therefore design a long chain selection algorithm and long chain interest encoder to obtain global-view long chain interest from the global click graph. We design a gated network to integrate long chain interest with neighbor interest to achieve the collaborative interest among similar users. Subsequently we aggregate it with local news category-enhanced representation to generate final user representation. Then candidate news representation can be formed to match user representation to achieve news recommendation. Experimental results on real-world datasets validate the effectiveness of our method to improve the performance of news recommendation\footnote{Code is in \url{https://github.com/EdwinWang37/GLoCIM}.}.

\end{abstract}



\section{Introduction}\label{introduction} 
News recommendation, vital for aligning content with users' reading preferences~\citep{das2007google}, requires accurate modeling of user interests given the daily production of news~\citep{DKN}.


In news recommendation, local modeling uses deep learning techniques like attention mechanisms, Convolutional Neural Network (CNN), and Long Short-Term Memory (LSTM) to extract semantic information from clicked news, generating user interest representations from local user click sequences for recommendations~\cite{NRMS}.


Methods based on local user interest have quickly reached a bottleneck in news recommendation, prompting the adoption of GNNs. These enhance recommendations by using external knowledge graphs to capture implicit interests. For instance, KIM~\citep{qi2021personalized} employs knowledge graphs to enrich representations of user history click news and candidate news.

\input{Figs/introlong.tex}


However, these methods inadequately capture common interest features for collaborative recommendations among similar users. Recent GNN research focuses on building local subgraphs from a global click graph based on all users' click sequences, as shown in Figure \ref{Fig:intro1-long}. GLORY~\citep{yang2023going} builds a global click graph from all users' click sequences to provide a global-view enhancing recommendations. Commonly, local graphs are constructed for individual users and analyzed with various GNNs~\cite{velickovic2017graph,chen2020simple,li2015gated}, but this approach incurs significant computational costs when modeling far-reaching linkage between two distant nodes in global click graph.


It is well understood that two key barriers of computational complexity problem obstruct the full exploitation of the global click graph. Firstly, the process of distilling information from the global graph, based on all users' click history, often yields a graph that is excessively large, typically incurring significant efficiency costs in terms of memory and computational speed. Secondly, GNNs~\cite{velickovic2017graph,chen2020simple,li2015gated}, which utilize subgraphs restrict short-range hop neighbors, exhibit limited expressive capabilities and face difficulties in capturing long chain interest. This limitation makes it challenging to effectively distill long chain interest from global click graph.


Inspired by metapath~\citep{fan2019metapath} from heterogeneous graph, we attempt to capture far-reaching linkage in the global click graph, thereby constructing the most valuable long chain for a user's each clicked news. We propose long chain selection algorithm and  interest encoder to distill long chain interest from the global click graph. A gated network integrates these with neighbor interest for a collaborative interest representation, which is merged with local category enhanced news interest to form the user representation. A candidate neighbor entity encoder then forms a candidate representation. Our main contributions are:

\begin{itemize}
\item  We design a long chain selection algorithm based on popularity, category and semantics to select long chains for each clicked news of all users in global click graph, so that we can select the most valuable long chains to better capture far-reaching linkage. 
\item We propose long chain attention network to distill the long chain interest at the news level and user level respectively, subsequently design a gated network to combine long chain interest and neighbor interest to enhance collaborative interest among users.
\item Extensive experiments on real world datasets demonstrate that GLoCIM improves news recommendation effectiveness, outperforming other baseline methods.
\end{itemize}

\section{Related Work}
A common approach is to study local news semantic representation through user's reading history with some deep learning techniques. Several approaches~\citep{DAE_RNN, LSTUR} have been refined to better model user interest based on their news interaction histories.  LSTUR~\citep{an2019neural} combines long-term interest represented by a user's ID with short interest derived from recent activities, while DAN~\citep{zhu2019dan} enhances user hidden sequential features of user’s clicks through attention-based parallel CNN. 
When the effectiveness of local semantic representation reaches its limitations, improving personalized recommendation can be accomplished by creating an extensive news graph that leverages the multi-dimensional data within it, User-as-graph~\citep{wu2021user} and DIGAT~\citep{mao2022digat} employ graph-based approaches to capture higher-dimensional information by leveraging user behavior patterns and dual graph interactions. KIM~\citep{qi2021personalized} uses a co-encoding method with a knowledge graph to understand the interactions between clicked and candidate news, enhancing the personalization of recommendations. HieRec~\citep{qi2021hierec} subsequently models user interest from subtopics to a comprehensive synthesis within a hierarchical interest framework. This method reflects a trend towards a more layered understanding of user interest in news recommendation.

Drawing on techniques that employ user-news graphs for news recommendation, it has been demonstrated that graph modeling significantly enhances the effectiveness of recommendation systems~\citep{graph-based-rec}. GERL~\citep{GERL} and GUND~\citep{GNUD} formulate news and users jointly in a bipartite graph to model news-user interaction. Some methods also incorporate a global click graph derived from user click history to enrich collaborative information. GNewsRec ~\citep{hu2020graph} introduces a heterogeneous graph model combined with GNNs and an attention-based LSTM to improve news recommendation. By integrating user, news, and topic interactions, the model distills both long-term and short-term user interest. GLORY~\citep{yang2023going} introduces a model that combines global and local insights for personalized recommendation. It uses a global-aware encoder to dive into a global click graph and gated GNN to enhance news representation via a historical aggregator. Nonetheless, these approaches have limitations in fully exploiting far-reaching linkage and optimizing the utilization of long chain interest among collaborative users, thus unintentionally or inadequately overlooking the long chain interest distillation.

\input{tables/comparison}

\section{Method}
GLoCIM is mainly divided into four phases: global click graph construction and news semantic representation generation, long chain and neighbor subgraph selection, interest distillation, and user-news matching, which is shown in Figure \ref{Fig:model}. 
We mainly propose long chain selection algorithm to dynamically obtain long chains, and then design long chain interest encoder to distill long chain interest and gated neural network to fuse long chain interest with neighbor interest in global click graph.

\noindent\textbf{Problem Formulation.} The click history sequence of a user $u$ can be denoted as $\mathbf{H}_u=[d_1, d_2, ..., d_H]$, where $H$ is the number of historical news articles. Each news article $d_i$ has a title, which contains a text sequence $\mathbf{T}_i=[w_1, w_2, ..., w_T]$ consisting of $T$ word tokens, and an entity sequence, which is denoted by $\mathbf{E}_i=[e_1, e_2, ..., e_E]$ consisting of $E$ entities. The objective is to predict the level of interest $s_{u,c}$ for a given candidate news article $d_c$ and user $u$, which reflects the likelihood of a clicking action occurring between them. 

\input{Figs/model.tex}

\subsection{Category-enhanced news Representation}
In the first phase of news semantic representation generation, to model user behaviors to filter the most valuable long chains in global click graph, we leverage category information to enhance news representation. We limit our analysis to the last $L_{his}$ clicked news in user $u$'s history $\mathbf{H_u}$ and the candidate news $d_c$. Only the first $L_{title}$ words of each news title are considered for input.

\noindent\textbf{News Encoder.} We initialize each news $n$ with a word embedding layer using GloVe~\citep{glove} embeddings to convert news text into embedding vectors $\mathbf{x}_n$. These are processed by a multi-head self-attention (MSA) layer to produce each news semantic representations $\mathbf{X}_n$:
\begin{equation}
    \mathbf{X}_n = \text{MSA}(\mathbf{x}_n)
\end{equation}



Thus, the title representation $\mathbf{X}_n$ is enhanced with news category $c_n$ and subcategory $sc_n$ using concatenation and trainable $W_{\text{mix}}$ and $b_{\text{mix}}$ to generate news representation $\mathbf{C}_n$. We can integrate both sementics and category to make preparations for better matching between clicked news:
\begin{equation}
    \mathbf{C}_n =  
    (\mathbf{X}_n \oplus c_n \oplus sc_n ) \cdot W_{\text{mix}} 
    + b_{\text{mix}}
\end{equation}

\noindent\textbf{Local News encoder.} In the third phase of semantic interest distillation, a text attention layer aggregates news final representations to form user local interest news representation $\mathbf{h}^{ln}$ as follow:
\begin{equation}\label{eq: attn1}
    \mathbf{h}^{ln} = \sum_{i=1}^{L_{his}} \alpha_i \mathbf{C}_i
\end{equation}
\begin{equation}
        \alpha_i = \frac{exp(\mathbf{q}^\top \tanh(\textbf{W} \cdot \mathbf{C}_n^i))}{\sum_{j=1}^{L_{his}} exp(\mathbf{q^\top} \tanh(\textbf{W} \cdot  \mathbf{C}_n^j))}
\end{equation}
where $\mathbf{q}$ is a query vector, $\mathbf{W}$ is a trainable matrix.


We feed pre-trained TransE embeddings of entities in news into an entity self-attention network to form entity representation $\mathbf{X}_e$, and an attention network aggregates these into user local interest entity representation $\mathbf{h}^{le}$.

\subsection{Long chain Interest modeling}
To model long chain interest, we first select the most valuable long chains through long chain selection algorithm, and then use the long chain interest encoder to distill interest. After every certain number of local news encoder training steps in the above process, we use updated news representations to select updated news long chains and distill long chain interest, thereby modeling the final long chain interest. The details are shown below:

\subsubsection{Long chain Selection algorithm}
The long chain selection algorithm consists of alternative long chain selection and pruning from global click graph \(G\) as shown in algorithm \ref{alg:1}. \(G\) represents the relationships between news articles, constructed from the temporal sequence of click histories of all users. Since each user possesses its own click histories, \(G\) is subsequently constructed to facilitate recommendations.

\noindent\textbf{Alternative long chain Selection.} For a user \(u\) within the range from \(u_1\) to \(u_N\), a sequence \( (n_A, n_B, n_C, \ldots) \) is formed in click timestamp sequence. For any two adjacent news article in the sequence, a directed pair \((n_A, n_B)\) can be represented within a global click graph. For each directed pair \((n_A, n_B)\), when the click sequence of user \(u\) \( (n_{1,u}, n_{2,u}, ... , n_{i,u}, n_{i+1,u} \ldots) \) satisfies $n_{i,u} = n_A, n_{i+1,u} = n_B$, the click frequency of user \(u\) can be defined as \(f_u(n_A \rightarrow n_B) = 1\); otherwise, \(f_u(n_A \rightarrow n_B) = 0\).

In addition with semantics and category, our third long chain selecting strategy utilizes the temporal popularity of news clicks, making news directed pairs with more clicks more likely to be captured. Therefore, to accurately utilize the collaborative relationships among different users to match user interest, for the global click graph, different users may have a non-zero click frequency for the same directed pair. Thus, an accumulated frequency for a directed pair can be defined:
\begin{equation}
af(n_A \rightarrow n_B) = \sum_{i=u_1}^{u_N} f_i(n_A \rightarrow n_B)
\end{equation}

We apply a similarity-based method to dynamically find the most relevant neighbors for each news article in \(G\). For a target news \( n_t \), \(\forall n \in G, af(n \to n_t) \neq 0\), \( n \) is a neighbor of \( n_t \).

Define a set \( N_{n_t}^1 = \{ n \in G \mid af(n \rightarrow n_t) \neq 0 \} \) as the first hop neighbor set of \( n_t \). Subsequently, we define \( N_{n_t}^{i+1} = \{ n_{t,i} \in N_{n_t}^i, n \in G \mid af(n \rightarrow n_{t,i}) \neq 0 \} \) as the i+1 hop neighbor set of \( n_t \). However, as the hop count \( i \) increases, the size of \( N_{n_t}^{i+1} \) grows exponentially. Therefore, we select an alternative long chain (ALC\(_i\)) containing a small number of important neighbors from each hop's neighbor set. We select the alternative long chain with branch \(m\) using top \(m\)  click weight news in \(N_{n_t}^1\). Here, the click weight is defined below:
\begin{equation}
W_j^1 = \frac{af(n_j \to n_t)}{\sum_{k = 1}^{|N_{n_t}^1|} af(n_k \to n_t)}, n_j \in N_{n_t}^1
\end{equation}


\begin{algorithm}
    \renewcommand{\algorithmicrequire}{\textbf{Input:}}
    \renewcommand{\algorithmicensure}{\textbf{Output:}}
    \caption{Long Chain Selection}
    \label{alg:1}
    \begin{algorithmic}[1]
    \REQUIRE  Global click graph $G$, Target news $n_t$, Long chain length $len$, Alternative long chain branches $m$, Total training step $st$, The weight set of neighbor news at $i$ hop is $W^i$, Total update times is $k$.
    \ENSURE   Long chain for $n_t$:$LC(n_1, n_2, \dots, n_{len})$
    \STATE $LC \leftarrow \text{Queue}()$ 
    \WHILE{$\left(\text{step} \bmod \frac{st}{k}\right) = 0$}
        \STATE ${\forall{n}} \in G, \text{ Local News Encoder}(n_t) \rightarrow C_{n_t} $
        \STATE $ALC_i$'s hop is $i, i \leftarrow 0$
        \STATE $n_c \leftarrow n_t$
        \WHILE{$ i < len$}
            \STATE $i++$
            \STATE $W^i \leftarrow caculate(N_{n_c}^1)$ 
            \STATE sort($W^i,  \text{ } descending$) 
            \STATE $ALC_i \leftarrow W^i[0:m-1]$
            \STATE Find $n_c'$ in $ALC_i$ by
            \Statex $\text{    }\quad \arg\max_{s \in ALC_i} \cos(C_{n_t}, C_{n_s})$
            \STATE $n_c \leftarrow n_c'$
            \STATE $LC.\text{enqueue}(n_c)$ 
        \ENDWHILE
    \ENDWHILE
    \end{algorithmic}  
\end{algorithm}




\noindent\textbf{Alternative long chain  Pruning. }
Then we designed alternative long chain pruning to select the most important news among the \( i \)-th hop neighbors of \( n_t \) from \( ALC_i \) to add to the long chain, as the \( i \)-th news article for \( n_t \)'s long chain. We calculate \(\text{cos similarity}(C_{n_t}, C_{n_s})\) for each \(s\) in \(\text{ALC}_i\) and select the most similar article \(s^*\) to add to the tail of long chain for \(n_t\). Thus the long chain  is refined  based on semantics, category, and popularity to  enhance the computation of similarities.
\begin{equation}
s^* = \arg \max_{s \in ALC_i} \cos({C_{n_t}, C_{n_s}})
\end{equation}

We iterate alternative long chain selection and pruning in different hop neighbor sets of \( n_t \). Thus in each hop of \( n_t \)'s neighbor, we compress the exponentially large neighbor set \( N_{n_t}^{i+1} \) to a constant size \( O(1) \).

In summary, we randomly initialize parameters of local news encoder for an initial long chain selection, and then periodically reselect after a certain number of training steps. During the training news representation process, we gradually extract semantic, category, and popularity information to enhance selection ability. This dynamic process ensures valuable long chains are identified during each reselection as shown in algorithm~\ref{alg:1}.


\subsubsection{Long chain Interest Encoder}

To effectively capture far-reaching linkage, we propose a long chain attention network to distill long chain interest.

At the news level, we now have the most valuable long chain for every clicked news, the correlation between different nodes in one long chain is typically similar news with a chronological progression. For instance, interest in football can be identified from several non-adjacent football news clicks in this long chain. Thus, we propose to apply news-level multi-head self-attention to enhance the the long chain interest representations $\mathbf{l}_{n}$  of news by capturing these non-adjacent interest. The corresponding formula is as follows:
\begin{equation}\label{eq: attn2}
    \mathbf{l}_{n} = \sum_{i=1}^{len} \beta_i C_i \quad 
\end{equation}
\begin{equation}
    \beta_i = \frac{exp(\mathbf{q}^\top \tanh(\textbf{W}_1 \cdot \mathbf{C}_n^i))}{\sum_{j=1}^{len} exp(\mathbf{q^\top} \tanh(\textbf{W}_1 \cdot  \mathbf{C}_n^j))}
\end{equation}
where $\mathbf{q}$ is a query vector, $\mathbf{W}_1$ is a trainable matrix, ${C}_n^i$ and ${C}_n^j$ represent the category enhanced news representation by local news encoder, $\beta_i$ represents the attention weight of a specific hop ${C}_n^i$ in the long chain of the current news learned through multi-head self-attention, $\mathbf{l}_{n}$ represents the long chain representation of the current clicked news.

Different long chain may have different informativeness in representing users. Therefore, we personalize user interest on these long chains. At the user level, we propose to apply an additive attention network to select important long chain interest to learn more informative user representations. This allows each user to form its own specific long chain interest representation $\mathbf{L}_{u}$ based on $\mathbf{H_u}$.
\begin{equation}\label{eq: attn}
    \mathbf{L}_{u} = \sum_{i=1}^{L_{his}} \gamma_i \mathbf{l}_{n}^i \quad 
\end{equation}
\begin{equation}    
    \gamma_i = \frac{exp(\mathbf{q}^\top \tanh(\textbf{W}_2 \cdot \mathbf{l}_{n}^i))}{\sum_{j=1}^{L_{his}} exp(\mathbf{q^\top} \tanh(\textbf{W}_2 \cdot  \mathbf{l}_{n}^j))}
\end{equation}
where $\mathbf{q}$ is a query vector, $\mathbf{W}_2$ is a trainable matrix. $\gamma_i$ represents the attention weight of the long chain representation $\mathbf{l}_{n}$ of a clicked news article for the current user that we have learned.

\subsection{Neighbor Interest Encoder}

In the third phase of neighbor interest distillation, for each user click history, we construct a subgraph from the global click graph to analyze specific user interest. For each news in $\mathbf{H_u}$ of user $u$, we select the top $M_n$ neighbors from multiple hops in the global click graph based on popularity in global click graph. We employ the gated GNN, which integrates a gated recurrent unit \cite{GRU}, to encode the neighbor news embedding:
\begin{equation}
    \mathbf{N}_n = \text{GGNN}(\mathbf{h}^{ln}_n)
    \label{eq: ge}
\end{equation}
\begin{equation}
    \mathbf{h}_i^{(l+1)} = \text{GRU}\left(\sum_{j \in \mathcal{N}(i)} W_g \cdot \mathbf{h}_j^{(l)}, \mathbf{h}_i^{(l)}\right)  
\end{equation}
where $\mathbf{h}^{(0)} = \mathbf{h}^{ln}$, initializing with  $\mathbf{h}^{ln}$. $\mathcal{N}(i)$ represents the neighbors of node $i$, and $W_g$ is a trainable matrix.

The neighbor aggregator, following \cite{wu-etal-2019-neural}, utilizes a multi-head attention layer and an attention pooling layer to aggregate $\mathbf{O_u}$ to derive the user final long chain to enhance neighbor representation $\textbf{LN}_u$ , akin to Eq.~\ref{eq: attn1}.

\subsection{Collaborative Interest Fusion and user representation}
Considering the complexity of long chain interest and neighbor interest, we initially exclude using direct operations like multiplication, concatenation or addition. This is because these two interest are not entirely independent and often occasionally overlap. Therefore, we design a gated fusion network based on LSTM to balance long chain interest and neighbor interest with a common ${G}_u $ to separately affect long chain and neighbor, thus model multi-level interest by integrating long chain interest ${L}_u$ with current neighbor interest ${N}_u$:
 \begin{equation}
    \bm{G}_u = \mbox{sigmoid}(\bm{W}^1 \bm{N}_u + \bm{W}^2 \bm{L}_u + b)
    \end{equation}
    Thus we can fuse long chain and neighbor interest to model collaborative interest $\mathbf{LN}_u$:
    \begin{equation}
    \mathbf{LN}_u = (\bm{1} - \bm{G}_u) \odot \bm{N}_u + \bm{G}_u \odot \bm{L}_u
\end{equation}
where $\odot$ is element-wise multiplication, $\mathbf{{W}^{1}}$ and $\mathbf{{W}^{2}}$ are trainable matrices, $b$ is a bias of fusion.

As for global-local aggregator, for each user, we derive the local news representation $\mathbf{h}^{ln}$, the local entity representation $\mathbf{h}^{le}$, and the long chain enhanced neighbor representation $\mathbf{LN}_u$. We employ an attention pooling network to combine these three representations into a user representation $\textbf{emb}_{user}$.

\subsection{Candidate News Modeling and Prediction}

This candidate encoder leverages a global entity graph~\citep{yang2023going} for candidate news representation. The graph is constructed from entity occurrences in news. Entity encoding begins with selecting the top \( M_e \) neighbors for each entity in candidate news \( d_c \). The local-view entity encoder's embedding layer is then used to derive \( \mathbf{x}_c \), which is refined by \( MSA(\mathbf{x}_c) \) to yield the global entity representation \( \mathbf{h}_c^{ge} \). The candidate news representation \( \mathbf{emb}_{cand} \) is synthesized by pooling the local news \( \mathbf{h}_c^{ln} \), local entity \( \mathbf{h}_c^{le} \), and global entity \( \mathbf{h}_c^{ge} \) representations.

We calculate the click score for each news article by taking the inner product of the user embedding $\mathbf{emb}_{user}$ and the candidate embedding $\mathbf{emb}_{cand}$. We use negative sampling \cite{NPA} in training. In each session of the training set, we use one positive and \( K \)
 negative samples, whose click probability score is \( \hat{y}^+ \)  and \( \left[ \hat{y}_1, \hat{y}_2, \dots, \hat{y}_K \right] \) respectively. We then optimize the log-likelihood loss $\mathcal{L}_{NCE}$ for positive samples during training:

\begin{equation}\label{eq13}
    \mathcal{L}=-\sum_{i=1}^{|\mathcal{D}|}{\log\left(\frac{e^{\hat{y}^+_i}}{e^{\hat{y}^+_i} + \sum_{j=1}^{K}{e^{\hat{y}^-_{i,j}}}}\right)}
\end{equation}

\input{tables/main_experiments}
\section{Experiment Setup}

\subsection{Dataset and Experiment Settings}
Mind~\citep{MIND} is a real-world news recommendation corpus including Mind-small and Mind-large datasets, as shown in table \ref{table:datasets}. so we conduct our experiments on them.
Following the previous work \cite{NPA}, each user’s historical record is constrained to include no more than $L_{his} = 50$ of the most recent news articles, with the length of each article’s title not exceeding $L_{title}=30$ words. We utilize up to $L_{entity}$ = 5 entities per news article, adjacency entities involve 10 neighbor entities. The news neighbor interest encoder considers 2 hops with 10 neighbor news each hop. We employ a 2-layer GGNN~\citep{li2015gated} for neighbor interest encoder. We initialize with 300-dimensional GloVe and 100-dimensional TransE embeddings from MIND dataset, configuring news representations at 400 dimensions. $m$ is set to 3 to select top 3 ALC neighbors at each hop, while $len$ is set to 8 for maximum length of a long chain. Adam \cite{Adam} optimizes our model with a learning rate of $2e^{-4}$, incorporating a 10\% warm-up and linear decay. The negative sampling rate is set at $K_{neg} = 4$, with all settings validated against a test set.

\subsection{Comparison Methods}
We select four groups of baselines to compare:
\textbf{Sequence-based methods}: 
\textbf{(1)} \textbf{LSTUR}~\cite{LSTUR} combines GRU-based neighbor user interest embedding with user ID embedding for dynamic user representation.
\textbf{(2)} \textbf{DAN}~\cite{zhu2019dan} integrates LSTM attention and candidate awareness to dynamically model user interest from interaction history.
\textbf{(3)} \textbf{CUPMAR}~\cite{tran2021deep} uses GRU to understand user preferences from reading history, employing a news encoder for article properties and a user-profile encoder for user context interest.

\textbf{Attention-based methods}: 
\textbf{(1)} \textbf{NRMS}~\cite{NRMS} employs multi-head self-attention networks to derive representations for news and users.
\textbf{(2)} \textbf{HieRec}~\cite{qi2021hierec} utilizes a hierarchical interest tree to represent and match user interest at multiple granularities.

\textbf{GNN-based methods}: 
\textbf{(1)} \textbf{User-as-Graph}~\cite{wu2021user} employs heterogeneous graph and graph pooling to depict complex user-action relationships.
\textbf{(2)} \textbf{KIM}~\cite{qi2021personalized} models user-news interactions using a co-encoder and a knowledge graph for user interest representation.
\textbf{(3)} \textbf{DIGAT}~\cite{mao2022digat} utilizes a dual graph attention network to align news and user graph channels for effective recommendation.

\textbf{Global-view click graph methods}: 
\textbf{(1)} \textbf{GNewsRec}~\cite{hu2020graph} merges user and news graph data with topic categorization into a unified GNN model for enriched news representation.
\textbf{(2)} \textbf{GLORY}~\cite{yang2023going} combines a global click graph with a gated GNN to create a hybrid news encoder that enhance global representation.

\section{Experiment Results and Analysis}


\subsection{Main Comparison Results} 
Table~\ref{table:main_experiments} presents the performance comparison results, with GLoCIM consistently achieving the highest across all metrics. We select three sets of experiments for comparison.


First, on the Mind-small dataset, HieRec~\citep{qi2021hierec} outperforms DIGAT~\citep{mao2022digat} by 0.17 in AUC, highlighting local user click history attention benefits in a smaller dataset.

Next, on Mind-large dataset, User-as-Graph outperforms HieRec by 0.2 in AUC, showing that GNN-based methods with higher-dimensional information excel over those based on user local click history attention in larger dataset.

Finally, on the same dataset, GLORY~\citep{yang2023going} achieves substantial gains in nDCG@5 and nDCG@10, increasing by 0.32 and 0.65 respectively, emphasizing the superiority of global click graph methods in enhancing ranking effectiveness.

Compared to the three aforementioned suboptimal methods, our method shows a clear advantage. In the case of Mind-small dataset where HieRec holds an advantage, GLoCIM's integration of long chain interest, combined with the use of the global click graph under data scarcity, enables it to outperform HieRec by a margin of 0.26 in AUC. On Mind-large dataset, GLoCIM also excels, surpassing the suboptimal User-as-Graph by 0.29 in AUC and GLORY by 0.36 in nDCG@5 and 0.39 in nDCG@10. This performance improvement comes from the fact that GLoCIM not only utilizes the neighbor interest but also distills long chain interest in the global click graph to capture far-reaching linkage and effectively fuse them to enhance user representation.
\input{Figs/ab-tuu}

\subsection{Ablation Analysis} 
\label{CNE_abltions} 
\input{tables/hyperparameter}

\input{tables/network_comparison} To verify the validity of our fundamental modules, we conducted four sets of experiments to dive into the effectiveness of the Long chain Interest Encoder (LE), Long chain Selection Algorithm (LSA), and Collaborative Interest Fusion Module (CF), with our base module being the Neighbor Interest Encoder (NE). We conducted four experiments: (1) \textbf{NE}: only use neighbor interest encoder (2) \textbf{NE+LE}: incorporate Long chain Interest Encoder but select long chain only by popularity (3) \textbf{NE+LE+LSA}: incorporate LSA but the fuse module is only element-wise multiplication (4) \textbf{NE+LE+LSA+CF(GLoCIM)}: full method.

All other settings remain the same as the full method, with only one component being replaced or removed. We evaluated these experiments on MIND-small dataset. The experimental results are shown in Figure~\ref{Fig:ab-tuu}. The full method performs the best, but the individual components also play a critical role in achieving optimal results. The role of LSA is the most apparent, proving that LSA selects the most valuable long chains, thereby effectively extracting and utilizing the long chain information in the global click graph.

\subsection{Analysis on Hyperparameters}

We explore the hyperparameter of hop number $len$ in a long chain to assess its impact on performance. Experimental results from MIND-small dataset, as illustrated in Figure ~\ref{hyper}, indicate that GLoCIM attains optimal recommendation performance when the number of hops is set to 8. This finding suggests that a lower number of hops may be inadequate for effectively distilling essential long chain interest, while a higher number may introduce excessive noise, thereby degrading the performance.

\subsection{Impact of Graph Encoder}

Far-reaching linkage modeling in global click graph faces massive computational overhead. So we evaluate impact of graph encoder on MIND-small dataset as shown in Table \ref{table:network_comparison}.

In order to extract far-reaching linkage, we adopt a common approach to compare: extracting local subgraphs from the global click graph to capture farther nodes, i.e., expanding the number of node hops. We use three popular GNNs (GAT, GCN, GGNN) to model the subgraphs.  Observations show that as hops increase, training, inferring time, and overhead grow exponentially, while performance declines, indicating traditional methods' inefficiency in capturing far-reaching linkage.

Subsequently, our findings show that GLoCIM's AUC surpasses that of the best-performing GGNN-2 by over 0.53, indicating a more thorough exploitation of the global click graph and a better distillation of long chain interest that are often overlooked, while maintaining similar inference time and storage costs. It is important to note that our training time exceeds that of the optimal GGNN-2, which is reasonable given the time costs associated with the long chain selection algorithm. However, considering that the update period in the recommendation real-world scenario are not frequent, we believe that investing in longer training time to achieve higher accuracy is acceptable.

\section{Conclusion}

This paper presents GLoCIM, a novel news recommendation method that modeling a global-view long chain interest. We first design alternative long chain  selection with news click popularity, then prune the alternative long chain  based on enhanced news semantics and category representation. GLoCIM incorporates a neighbor and long chain interest encoder, distilling long chain  and neighbor interest to enhance global click graph information extraction. Tests on two real-world datasets demonstrate that GLoCIM surpasses baseline methods.

\section{Limitations}

Training GLoCIM on static public news datasets with restricted time frames, which don't reflect the dynamic nature of real-world data. As a result, it may struggle to adapt to real-time changes in news recommendation, potentially impairing its generalization and robustness in evolving environments.

\bibliography{anthology,custom}

\appendix
\input{Figs/casestudy.tex}

\section{Loop chain. } Long chain selection inherently includes recursive loops, such as \( n_A \rightleftharpoons n_B \), resulting in sequences like \( n_A \rightarrow n_B \rightarrow n_A \rightarrow n_B \).  While avoiding loops could reduce redundancy, it may also limit the distillation of long chain interest. Therefore, we intentionally allow such cycles.

\section{Datasets}
\input{tables/dataset}

\section{Case Study} 
We conduct a case study to evaluate GLoCIM and compare it with GLORY, which is currently the best on the global click graph, as shown in Figure \ref{Fig:case}.

We display one clicked news article from a randomly chosen user in the same impression. We show the news suggested by GLoCIM and GLORY, where only one candidate news "Fan Poll" was clicked. We also present the selected representative long chain related to this clicked news. Several observations can be obtained from this.

Firstly, our long chain can select candidate news that are closely related to the news in the global click graph. our long chain selection includes "NFL Week 7" and "Le'Veon Bell's knee injury", rather than news related to the finance domain such as "Mark Hurd" and "Waterfront City". We can clearly see our selection mechanism from the Figure \ref{Fig:case}. First, we calculate the corresponding weights based on accumulated frequency to obtain the alternative long chain: "Iranian students", "Mark Hurd" and "NFL Week 7". Then, by combining semantic information and category information to obtain the category-enhanced representation, we calculate the cosine similarity and finally add the highest score "NFL Week 7" to our long chain. Repeating the above process, we can obtain the most valuable long chain. Therefore, although finance-related news appears twice in the first hop, it is still not considered as our candidate long chain.

Secondly, GLoCIM ranks the candidate news clicked by the user higher than GLORY. We can observe that the current user actually clicked on "Fan Poll" in a real-world scenario, with GLoCIM recommending a rugby-related news article first, whereas GLORY first recommended "Katie Hill", a finance-related news article. This occurs as GLORY only uses GGNN to model user interest in neighbor hops. Given that the neighbor news includes 'Mark Hurd' and 'waterfront city', GGNN over-learns news in the financial domain. Therefore, GLORY tends to give finance related news a higher ranking. In contrast, we can not only capture neighbor news but also model relevant news representations in the most valuable long chain, such as the fifth hop news "Bell's knee injury" , thereby enhancing the recommendation accuracy.

Different from GLORY, GLoCIM uses a long chain encoder and collaborative interest fusion, which can better distill long-chain interest in the global click graph than GLORY.

\end{document}

%% file: Figs/introlong.tex
\begin{figure}[t]
\centering
\includegraphics[width=1\columnwidth]{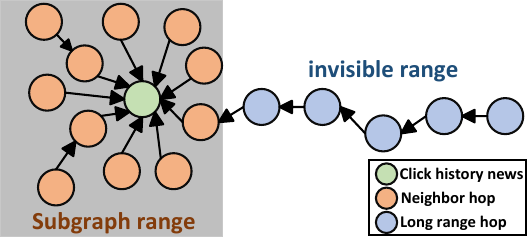}

\captionsetup{font=10pt}
\caption{
 An example of some news click global graph with traditional limited hop modeling without long chain selection and utilization.
}
\label{Fig:intro1-long}
\end{figure}

%% file: tables/comparison.tex



\begin{table}[htbp]
\caption{Comparison of methods based on various analytical dimensions: local modeling (C1), local graph assistance (C2), global graph utilization (C3), and global long chain interest distillation (C4).}
\small
\setlength\tabcolsep{3pt}
\begin{tabular}{p{110pt}cccc}
\toprule
Method & \textbf{(C1)} & \textbf{(C2)} & \textbf{(C3)} & \textbf{(C4)} \\
\midrule 
LSTUR~\cite{LSTUR}, CUPMAR~\cite{tran2021deep}, DAN~\cite{zhu2019dan}, NRMS~\cite{NRMS} , HieRec~\cite{qi2021hierec}& $\boldsymbol{\checkmark}$ & - & - & - \\
\midrule 
KIM~\cite{qi2021personalized}, DIGAT~\cite{mao2022digat}, User-as-Graph~\cite{wu2021user} & $\boldsymbol{\checkmark}$ & $\boldsymbol{\checkmark}$ & - & - \\
\midrule 
GLORY~\cite{yang2023going}, GNewsRec~\cite{hu2020graph} & $\boldsymbol{\checkmark}$ & $\boldsymbol{\checkmark}$ & $\boldsymbol{\checkmark}$ & - \\
\midrule 
GLoCIM & $\boldsymbol{\checkmark}$ & $\boldsymbol{\checkmark}$ & $\boldsymbol{\checkmark}$ & $\boldsymbol{\checkmark}$ \\
\bottomrule
\end{tabular}
\label{table:existing_algorithms}
\end{table}

%% file: Figs/model.tex
\begin{figure*}[t]
\centering
\includegraphics[width=\textwidth]{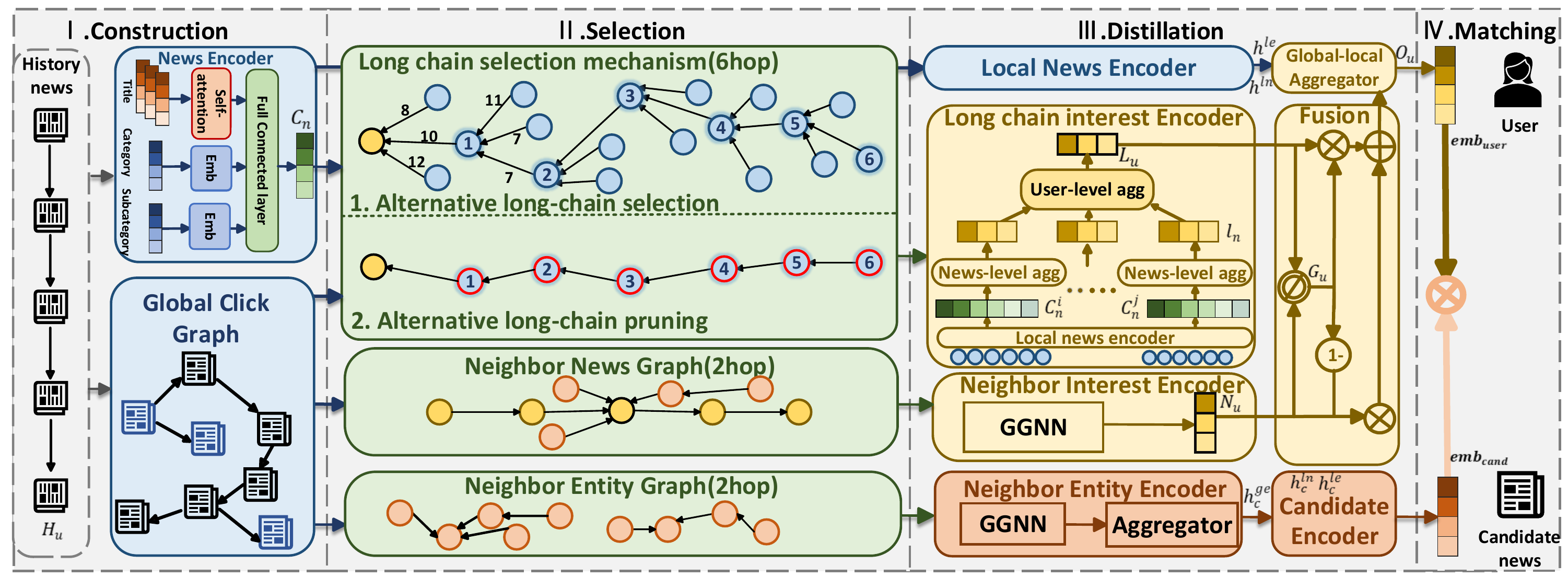}
\captionsetup{font=10pt}
\caption{
The overall architecture of our model. The graph construction is based on the user click history.
}
\label{Fig:model}
\end{figure*}

%% file: tables/main_experiments.tex
\begin{table*}[t]
\centering
\caption{Main results on MIND datasets. \textbf{Bold} means the best performance of baseline methods, while \underline{\textbf{underline}} means GLoCIM performance, which is also the best in all metrics.  We repeated these experiments with $^\dagger$ three times and reported the average result. We outperform the best method at level p < 0.05.}
\label{table:main_experiments}
\resizebox{\textwidth}{!}{
\fontsize{9}{11}\selectfont
\begin{tabular}{l|cccc|cccc}
\hline
& \multicolumn{4}{c|}{\textit{MIND-small}} & \multicolumn{4}{c}{\textit{MIND-large}} \\ 
\hline
Method & AUC & MRR & nDCG@5 & nDCG@10 & AUC & MRR & nDCG@5 & nDCG@10 \\ 
\hline
CUPMAR & 64.15 & 29.61 & 32.89 & 39.02 & 65.82 & 31.23 & 34.35 & 40.84 \\
LSTUR & 65.87 & 30.78 & 35.15 & 40.15 & 67.08 & 32.36 & 35.15 & 40.93 \\
DAN & 65.14 & 30.04 & 32.98 & 39.52 & 66.88 & 32.81 & 35.79 & 41.91 \\
\hline
NRMS & 65.36 & 30.02 & 33.11 & 39.61 & 67.01 & 31.85 & 35.34 & 41.75 \\
HieRec & \textbf{67.95} & \textbf{32.87} & \textbf{36.36} & \textbf{42.53} & 69.03 & 33.89 & 37.08 & 43.01 \\
\hline
User-as-Graph & \textendash & \textendash & \textendash & \textendash & \textbf{69.23} & \textbf{34.14} & 37.21 & 43.04 \\
KIM $^\dagger$ & 67.07 & 31.83 & 35.23 & 41.58 & 68.45 & 33.74 & 36.76 & 42.47 \\
DIGAT w/o PLM & 67.82 & 32.65 & 36.25 & 42.49 & \textendash & \textendash & \textendash & \textendash \\
\hline
GNewsRec & 65.54 & 30.27 & 33.29 & 39.80 & 68.15 & 33.45 & 36.43 & 42.10 \\
GLORY $^\dagger$ & 67.68 & 32.45 & 35.78 & 42.10 & 69.04 & 33.83 & \textbf{37.53} & \textbf{43.69} \\
\hline
GLoCIM $^\dagger$& \textbf{\underline{68.21}} & \textbf{\underline{33.02}} & \textbf{\underline{36.69}} & \textbf{\underline{42.78}} & \textbf{\underline{69.52}} & \textbf{\underline{34.34}} & \textbf{\underline{37.89}} & \textbf{\underline{44.08}} \\
\hline
\end{tabular}
}
\end{table*}

%% file: Figs/ab-tuu.tex
\begin{figure}[t]
\centering
\includegraphics[width=83mm]{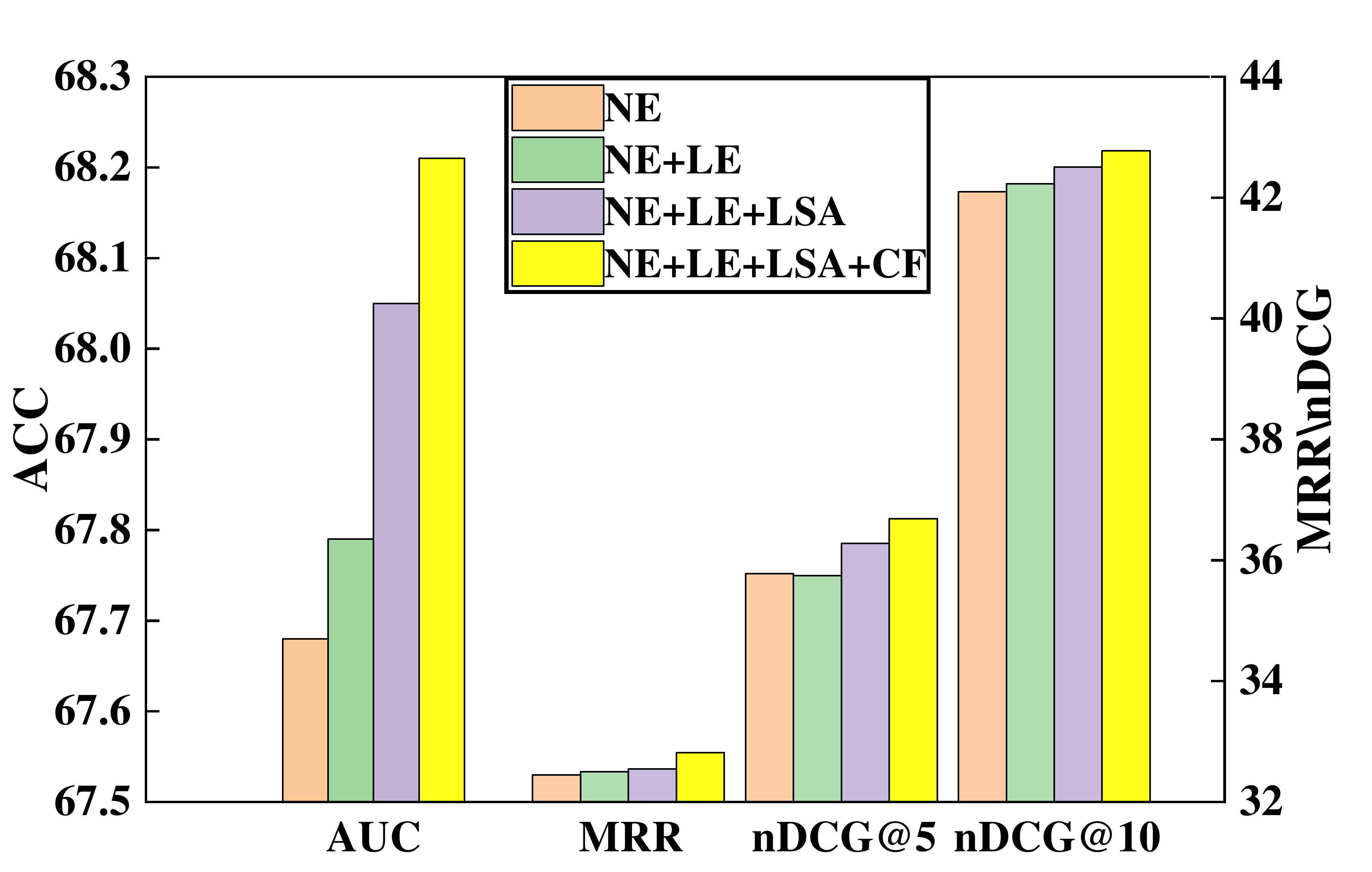}
\captionsetup{font=10pt}
\caption{
 Results of different module variants.
}
\label{Fig:ab-tuu}
\end{figure}

%% file: tables/hyperparameter.tex
\begin{figure}[t]
    \centering
    \begin{tikzpicture}
        \begin{axis}[
            width=0.45\textwidth,
            height=5cm,
            axis line style={-},
            axis x line*=bottom,
            axis y line*=left,
            ylabel={AUC},
            ylabel style={yshift=-0.3cm, font=\footnotesize\bfseries},  
            ytick distance=0.2,
            ymin=67.400, ymax=68.40,
            xtick= {2,4,6},
            xticklabels={6-hop, 8-hop, 10-hop},
            xticklabel style={align=center, font=\footnotesize\bfseries},
            xmin=1, xmax=7,
            every axis plot/.append style={line width=1.5pt},
            yticklabel style={font=\scriptsize\bfseries}
        ]
        \addplot [mark=*, color=mycolor5, mark options={scale=0.75}] coordinates {(2, 67.95) (4, 68.21) (6, 68.09)};
        \end{axis}

        \begin{axis}[
            width=0.45\textwidth,
            height=5cm,
            axis line style={-},
            axis x line=bottom,
            hide x axis,
            axis y line=right,
            ylabel={nDCG},
            ylabel style={yshift=0.3cm, font=\footnotesize\bfseries},  
            ytick distance=2.5,
            ymin=35.00, ymax=45.00,
            xtick= {2,4,6},
            xticklabels={1-hop, 2-hop, 3-hop},
            xmin=1, xmax=7,
            legend style={
            at={(0.5,0.4)},
            font=\scriptsize,
            anchor=north, legend cell align=left,legend columns=-1},
            every axis plot/.append style={line width=1.5pt},
            yticklabel style={font=\scriptsize\bfseries}
        ]
        \addplot [mark=diamond*, color=mycolor6] coordinates {(2, 36.19) (4, 36.69) (6, 36.32)};
        \addlegendentry{nDCG@5}
        
        \addplot [mark=square*, color=mycolor7, mark options={scale=0.75}] coordinates {(2, 42.23) (4, 42.78) (6, 42.38)};
        \addlegendentry{nDCG@10}
    
        \addlegendimage{mark=*, color=mycolor5}
        \addlegendentry{AUC}
        \end{axis}
    \end{tikzpicture}
    \caption{Impact of different hops}
    \label{hyper}
\end{figure}
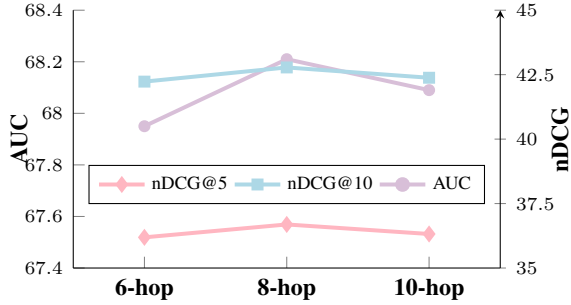

%% file: tables/network_comparison.tex
\renewcommand\cellalign{cc} 
\renewcommand\theadalign{cc} 
\renewcommand\theadfont{\bfseries} 
\renewcommand\theadgape{\Gape[0pt]} 
\begin{table}[t]
\centering
\caption{Model performance comparison. Average time cost of training/inferring 1000 samples.}
\label{table:network_comparison}
\fontsize{9}{11}\selectfont 
\begin{tabular}{c|c|c|c|c}
\hline
\textbf{Model} & \textbf{AUC} & \makecell{\textbf{Training}\\\textbf{Time (s)}} & \makecell{\textbf{Inferring}\\\textbf{Time (s)}} & \makecell{\textbf{GPU}\\\textbf{Mem (GB)}} \\
\hline
GGNN-1 & 67.35 & 10.62 & 11.80 & 6.84 \\
GGNN-2 & \textbf{67.68} & 21.02 & 21.02 & 13.60 \\
GGNN-3 & 67.49 & 121.91 & 55.84 & 20.43 \\
\hline
GAT-1  & 67.40 & 10.85 & 12.62 & 7.65 \\
GAT-2  & 67.60 & 21.78 & 23.55 & 14.37 \\
GAT-3  & 67.58 & 113.91 & 59.05 & 18.82
 \\
\hline
GCN-1  & 67.22 & 9.94 & 16.70 & 12.39 \\
GCN-2  & 67.45 & 21.01 & 15.09 & 17.32 \\
GCN-3  & 67.31 & 110.09 & 52.49 & 20.43
 \\
\hline
GLoCM-8   & \textbf{68.21} & 57.85 & 24.89 & 14.97 \\
\hline
\end{tabular}
\end{table}

%% file: Figs/casestudy.tex
\setlength{\fboxsep}{0pt}
\setlength{\fboxrule}{0pt}

\begin{figure*}[t]
\centering
\includegraphics[width=\textwidth]{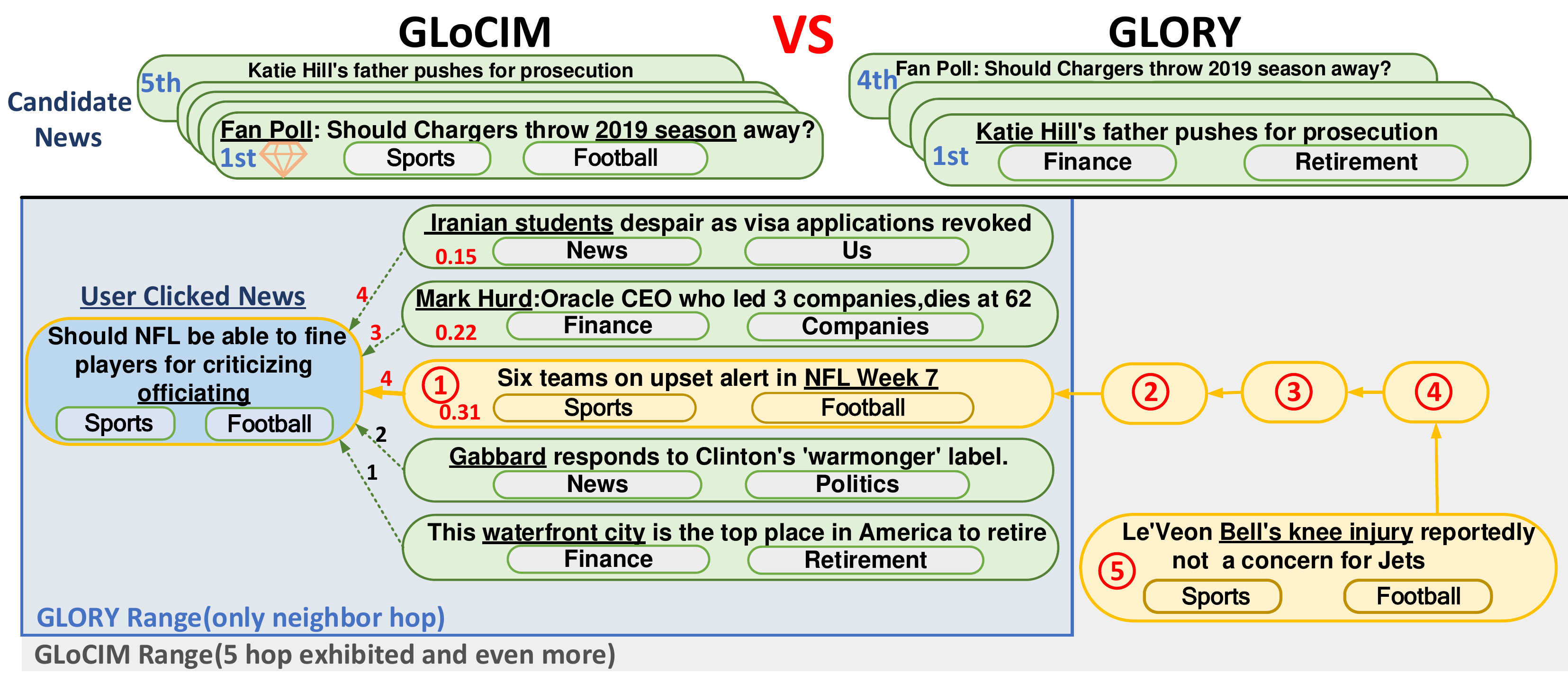} 
\captionsetup{font=10pt}
\caption{
Comparison of the effect of GloCIM and GLORY in utilizing global click graph for news recommendation. The circled red numbers represent different selected hop, and numbers on the arrows represent the accumulated frequency. The red numbers in the news represent the similarity to the user click news selected. The diamond icon represents the news that was actually clicked.}
\label{Fig:case}
\end{figure*}

%% file: tables/dataset.tex
\begin{table}[h]
\centering
\caption{Dataset Statistics.}
\begin{tabular}{lcc}
\toprule
 & MIND-small & MIND-large \\
\midrule
\# News & 65,238 & 161,013 \\
\# Categories & 18 & 20 \\
\# Sub-categories & 270 & 294 \\
\# Impressions & 230,117 & 15,777,377 \\
\# Clicks & 347,727 & 24,155,470 \\
\bottomrule
\label{table:datasets}
\end{tabular}
\end{table}